\documentclass[sigconf]{acmart}

\usepackage{wrapfig}
\usepackage{subcaption}
\usepackage{lipsum}
\AtBeginDocument{%
  \providecommand\BibTeX{{%
    Bib\TeX}}}

\copyrightyear{2024} 
\acmYear{2024} 
\setcopyright{acmlicensed}
\acmConference[DAC '24]{61st ACM/IEEE Design Automation Conference}{June 23--27, 2024}{San Francisco, CA, USA} 
\acmBooktitle{61st ACM/IEEE Design Automation Conference (DAC '24), June 23--27, 2024, San Francisco, CA, USA} 
\acmDOI{10.1145/3649329.3657391} 
\acmISBN{979-8-4007-0601-1/24/06}
\begin{document}

\title{Accel-NASBench: Sustainable Benchmarking for Accelerator-Aware NAS}

\author{Afzal Ahmad}
\email{afzal.ahmad@connect.ust.hk}
\orcid{0000-0003-4491-5440}
\affiliation{%
  \institution{The Hong Kong University of Science and Technology}
  \country{Hong Kong}
}

\author{Linfeng Du}
\email{linfeng.du@connect.ust.hk}
\orcid{0000-0002-3007-4890}
\affiliation{%
  \institution{The Hong Kong University of Science and Technology}
  \country{Hong Kong}
}

\author{Zhiyao Xie}
\email{eezhiyao@ust.hk}
\orcid{0000-0002-4442-592X}
\affiliation{%
  \institution{The Hong Kong University of Science and Technology}
  \country{Hong Kong}
}

\author{Wei Zhang}
\authornote{Corresponding author.}
\email{eeweiz@ust.hk}
\orcid{0000-0002-7622-6714}
\affiliation{%
  \institution{The Hong Kong University of Science and Technology}
  \country{Hong Kong}
}

\renewcommand{\shortauthors}{Ahmad et al.}

\begin{abstract}
One of the primary challenges impeding the progress of Neural Architecture Search (NAS) is its extensive reliance on exorbitant computational resources. NAS benchmarks aim to simulate runs of NAS experiments at zero cost, remediating the need for extensive compute. However, existing NAS benchmarks use synthetic datasets and model proxies that make simplified assumptions about the characteristics of these datasets and models, leading to unrealistic evaluations. We present a technique that allows searching for \textit{training proxies} that reduce the cost of benchmark construction by significant margins, making it possible to construct realistic NAS benchmarks for large-scale datasets. Using this technique, we construct an open-source bi-objective NAS benchmark for the ImageNet2012 dataset combined with the on-device performance of accelerators, including GPUs, TPUs, and FPGAs. Through extensive experimentation with various NAS optimizers and hardware platforms, we show that the benchmark is accurate and allows searching for state-of-the-art hardware-aware models at zero cost. 
\end{abstract}

\settopmatter{printacmref=false} % To remove ACM REFEREMCE FORMAT
\newcommand\blfootnote[1]{%
  \begingroup
  \renewcommand\thefootnote{}\footnote{#1}%
  \addtocounter{footnote}{-1}%
  \endgroup
}
\maketitle

\section{Introduction}
The proliferation of research in Neural Architecture Search (NAS), combined with the prohibitive costs of NAS evaluation, have highlighted the need for benchmarks and reproducibility~\citep{li2020random}. NAS benchmarks sidestep the expensive evaluation phase of NAS, which requires model training, by predicting accuracy using the architectural properties of the model, such as layer operations, filter sizes, and connectivity patterns. Since these benchmarks remove expensive model training from the evaluation phase of NAS, they are termed \textit{zero-cost}. Zero-cost NAS benchmarks using surrogate predictors have paved the way for cheap NAS evaluation while ensuring complete coverage of large search spaces at a minimal cost of benchmark construction~\citep{siems2020bench}. These benchmarks are constructed by evaluating a small but representative portion of the search space exhaustively and using these evaluations to train black-box surrogate models/predictors that can be utilized to obtain robust estimates of performance of unseen regions of the search space. However, currently available zero-cost NAS benchmarks only apply to small datasets such as CIFAR10/100~\citep{krishna2009cifar}, or cheap, synthetic variants of large datasets such as ImageNet16-120~\citep{chrabaszcz2017downsampled}.\blfootnote{Project partially funded by AI Chip Center for Emerging Smart Systems (ACCESS) and Hong Kong General Research Fund Grant Number GRF16213422. The authors also acknowledge Cloud TPU Support by Google TPU Research Cloud (TRC) program.}

Numerous works have pointed out the parameter-inefficiency and low quality of results obtained using \textit{proxy} datasets for search~\citep{cai2018proxylessnas}. Specifically, the process of searching on proxy datasets completely disregards the target dataset during the optimization process. Since the proxy and target datasets often have a massive disparity in complexity, the searched models exhibit a low parameter-efficiency. On the other hand, a few high-quality works that perform direct searches on large datasets have shown that a direct search yields models that exhibit higher parameter-efficiency~\citep{tan2019efficientnet, cai2018proxylessnas}. However, the high search cost of these methods, often in tens of thousands of GPU/TPU-hours~\citep{wan2020fbnetv2}, prohibits reproducibility.

On another front, the proliferation of high-performance hardware accelerators, combined with a drive for fast model serving, has given rise to accelerator-aware neural network design, intending to optimize multiple performance metrics, including accuracy and on-device throughput/latency. However, numerous research works have pointed out the inability of conventional model-specific complexity metrics, such as FLOPs and model size, to serve as accurate proxies for the model's on-device performance~\citep{gupta2020accelerator, li2021hw}. This is because performance is influenced by a range of device-specific factors such as memory bandwidth, data reuse, and the frequency/regularity of off-chip memory accesses. The device specificity of these factors makes the optimal choice of models targeting these devices contingent on the characteristics of the hardware being used.

In this work, we tackle the aforementioned challenges of unrealistic NAS evaluation by proposing a method that allows the construction of realistic NAS benchmarks at a fraction of the cost of existing methods. Using the proposed technique, we construct the first NAS benchmark for the ImageNet2012 dataset. Furthermore, for evaluation of multi-objective NAS methods, we also explore performance benchmarks for high-performance hardware accelerators, including GPUs, TPUs, and FPGAs. The contributions of this work are listed as follows\footnote{Project open-source at \url{https://github.com/afzalxo/Accel-NASBench}}

\begin{itemize}
    \item We propose a method to search for \textit{training proxies} that reduce the cost of model training for construction of NAS benchmarks for large-scale datasets by significant margins.
    \item Using the proposed technique, we construct the first NAS benchmark for the ImageNet2012 dataset, the de-facto large-scale dataset for visual recognition, while analyzing the impact of the training proxies on evaluation accuracy. 
    \item We also offer inference performance benchmarks for hardware accelerators, including Cloud TPUv2 and TPUv3, A100 and RTX-3090 GPUs, and Xilinx Zynq Ultrascale+ ZCU102 and Versal AI Core VCK190 FPGAs. 
    \item Through extensive experiments in both uni- and bi-objective search settings using various NAS optimizers and hardware platforms, we show that the constructed benchmark not only accurately simulates the real performance of models but also allows searching for state-of-the-art models at zero cost. 
\end{itemize}
\section{Preliminaries}
\subsection{Background}
NAS algorithms traverse a massive space of deep neural network models, called the search space, by evaluating selected models and using their performance and architecture specifications to guide the search towards better-performing solutions. NAS algorithms/optimizers decide the model to select and evaluate for traversing the search space in an efficient manner such that fewer evaluations can be performed to reach a good solution. The quality of solutions is often measured in terms of the model's accuracy on a dataset for uni-objective, and accuracy-latency/throughput (on-device performance) pair for bi-objective problems. Given that NAS algorithms search over massive spaces, ranging between $10^{10}$ - $10^{30}$ unique models, reaching an acceptable solution requires training and evaluation of thousands of models. This is an extremely computationally demanding process, with cost of running an optimizer per experiment ranging from GPU-hours to GPU-months. 

\begin{figure}[tp!]
\centering
% \vspace{-0.15in}
\includegraphics[width=0.88\linewidth]{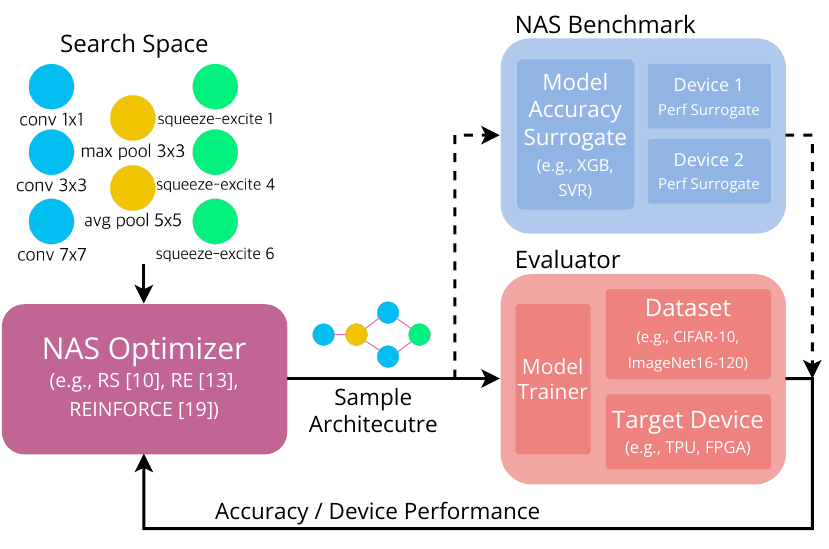}
\vspace{-0.10in}
% \captionsetup{format=plain,justification=justified}
\caption{NAS flow consists of an optimizer sampling architectures from a search space, followed by evaluation in terms of accuracy and on-device performance. A NAS benchmark sidesteps the expensive evaluation phase by using surrogate predictors that offer zero-cost evaluation.} \label{fig:nas-benchmark-flow}
\vspace{-0.20in}
\end{figure}
In recent years, NAS benchmarks have been proposed to allow zero-cost evaluation of optimizers, remediating the need for extensive compute resources and lowering the barrier of entry for researchers to evaluate new techniques. Fig.~\ref{fig:nas-benchmark-flow} shows the NAS flow consisting of an expensive evaluation step that is circumvented by the NAS benchmark. The most popular NAS benchmarks, called \textit{surrogate benchmarks}, use predictive models to estimate architecture accuracy/performance using their specifications, such as operation types, filter sizes, layer specifications, and connectivity patterns. These benchmarks are constructed by training and evaluating a small number of models from the NAS search space and using these evaluations to train surrogate predictors that can predict the accuracy/performance of unseen models in that space within a few milliseconds without model training and on-device measurements (hence zero-cost). The predictive models to use in benchmark construction are chosen based on their expressivity on the search space. Examples include gradient boosting such as XGB and LGB, support vector regression (SVR), and random forests.

\subsection{Challenges and Related Works}
While NAS benchmarks alleviate the computational burden of search by offering zero-cost evaluation, constructing NAS benchmarks is a computationally demanding process that requires thousands of model evaluations. The problem is exacerbated by increasing dataset and model sizes, for which training even a few hundred models is infeasible. In the following subsections, we highlight a few limitations of existing NAS benchmarks:

\subsubsection{Dataset Proxies and Sustainability.} Most NAS works utilize small datasets as \textit{proxies} for large-scale target datasets for search owing to the substantially lower search cost. A small number of research works that perform direct search on large-scale datasets have yielded models that beat state-of-the-art in parameter efficiency (e.g., ~\cite{tan2019efficientnet,howard2019searching} use reinforcement learning,~\cite{cai2018proxylessnas} used gradient-based search, while~\cite{real2019regularized} utilized regularized evolution directly on ImageNet2012). However, the construction of NAS benchmarks for large-scale datasets has remained elusive owing to the computational intractability of the evaluation of models on these datasets. As an example, NASBench-101~\cite{ying2019bench} utilized a staggering 100 TPU-years of compute to build a benchmark for CIFAR-10, largely considered to be a toy dataset. The benchmark covers a search space of only 423k models. Hence, while the utility of NAS benchmarks as an indispensable tool for democratizing NAS research cannot be ignored, the tremendous amount of compute for benchmark construction, combined with the limited coverage of search space and the use of the CIFAR-10 dataset, calls into question the sustainability and impact of NAS benchmarks~\cite{siems2020bench}. Recent NAS benchmarks use a downsampled variant of ImageNet2012, namely ImageNet16-120~\citep{chrabaszcz2017downsampled}, as a proxy for ImageNet2012~\cite{dong2020bench}. However, given the disparity in input size ($16\times16$ vs. $469\times387$ resolution), number of classes (120 vs. 1000), and number of samples in the dataset (0.15M vs. 12.4M images), ImageNet16-120 does not serve as a representative proxy for ImageNet2012. Our work aims to address this limitation by offering a technique for constructing benchmarks for large-scale datasets such as ImageNet2012 without using any dataset proxies.

\subsubsection{On-Accelerator Performance.} The rise of high-performance hardware accelerators has substantiated calls for accelerator-aware model search, aiming to optimize both model-specific metrics, such as accuracy, and device-specific metrics, such as throughput~\cite{gupta2020accelerator}. HW-NAS-Bench~\cite{li2021hw} was the first NAS benchmark that offered performance results for hardware platforms on NASBench-201~\cite{dong2020bench} and FBNet spaces; however, the benchmark suffers from unrealistic evaluation owing to the use of proxy datasets combined with analytical hardware performance results. The benchmark uses an unrealistic approximation of summing up the performance of unique blocks of the model rather than end-to-end performance measurements, leading to inaccurate results. Similarly, BRP-NAS~\cite{dudziak2020brp} offered latency measurements on the NASBench-201 space; however, the small size of the search space (only 16k models) combined with proxy datasets yield unreliable evaluation results. We use ImageNet2012, the de-facto large-scale dataset for visual recognition problems, combined with end-to-end throughput/latency measurements on various high-performance accelerators to offer results that are realistic and the resulting models deployable in real-world scenarios. 

%Numerous recent works have shown that model-specific complexity metrics such as FLOPs/model size serve as weak proxies for on-device performance~\cite{gupta2020accelerator, li2021hw}. This is due to the potential performance bottlenecks that could result from device-agnostic search, owing to architectural differences between hardware platforms, including different memory bandwidths, on-chip memory sizes, off-chip memory access latencies, parallelism, and pipelining capabilities. 
% For a more thorough review of existing NAS benchmarks and their construction methods, we invite the reader to explore~\cite{}.
%(For some cases, the coefficient of determination $R^2$ between measured and estimated performance is as low as $0.51$)
\section{Methodology}
In this section, we formulate our problem, aiming to reduce the benchmark construction costs for large-scale datasets. We then utilize our proposed technique to collect datasets for training and validating the surrogate predictors in an efficient manner.

\subsection{Search Space and Dataset}
To demonstrate our proposed techniques, we use a popular search space utilized by works that perform direct NAS using ImageNet2012, the MnasNet search space~\cite{tan2019mnasnet}. We do this to show that the benchmark allows zero-cost discovery of high-quality models that exceed models like EfficientNet-B0 in accuracy and on-device performance, a state-of-the-art model in this space. The lower bound on the true cost of search on this search space using the ImageNet2012 dataset is estimated to be over 91k GPU-hours (see~\cite{wan2020fbnetv2}). 

The MnasNet search space is a hierarchical block-based search space consisting of seven sequentially connected blocks/stages, each block hosting a certain number of mobile inverted bottleneck layers. Each block has a searchable kernel size $k$ and the number of layers $L$ in the stage. Squeeze-excitation $se$ is also searchable for each block. Expansion factor $e$ can take values in \{1, 4, 6\} while kernel size $k$ can take values in \{3, 5\} for each block. The number of layers $L$ in each block can take values in \{1, 2, 3\}. The search space holds roughly $10^{11}$ unique models. We utilize ImageNet2012 owing to its size and complexity, which yields generalizable models that can be deployed in real-world scenarios.

For experiments with additional search spaces and datasets for generalizability studies, please see our GitHub repository.

\begin{figure}[tp!]
\centering
% \vspace{-0.15in}
\includegraphics[width=0.85\linewidth]{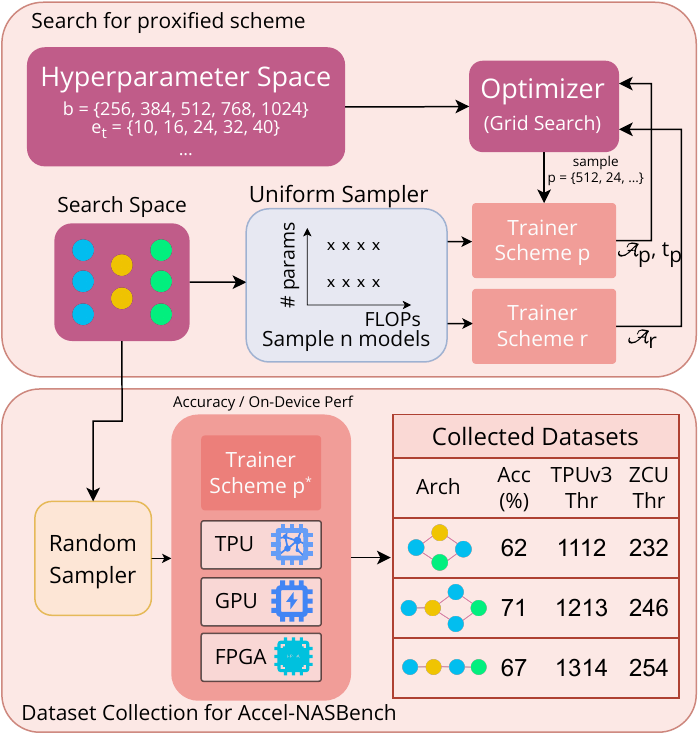}
\vspace{-0.1in}
% \captionsetup{format=plain,justification=justified}
\caption{(Top) Proposed method to search for proxified training scheme and (bottom) using the searched scheme ($p^{*}$) and hardware accelerators to collect datasets for construction of Accel-NASBench.} \label{fig:anb-methodology}
\vspace{-0.20in}
\end{figure}
\subsection{Search for Training Proxies}
The basis of our work is the fact that during the search phase of NAS, the optimizer aims to obtain architectures that are highly \textit{ranked} in terms of their performance rather than obtaining their true performance. Hence, approximations that lead to a reduction in the compute cost of evaluation are acceptable as long as they do not significantly impact the \textit{rankings} of architectures relative to their true ranks. One such approximation is \textit{training proxies}, utilized in hyperparameter optimization (HPO) for estimating model performance. A training proxy is a set of training hyperparameters that estimates the true training process while being computationally efficient. Examples of methods that use training proxies include successive halving and hyperband as they use the model's early-stage performance as a \textit{proxy} for true performance. 

Given the aforementioned observation, we formulate our problem as an architecture rank-correlation maximization problem. We aim to maximize the rank-correlation between true evaluation (i.e., using a reference, high-fidelity training scheme) and proxified evaluation (i.e., using a training scheme that uses approximations and sacrifices accuracy in favor of faster convergence), subject to a minimization constraint of training time under the proxified scheme:

\begin{equation} \label{eq:max_tau}
\begin{aligned}
    \max\ & \tau(\mathcal{A}_{p}, \mathcal{A}_{r}) \\
    \text{s.t.\ } & t_{p} \leq t_{\text{spec}}
\end{aligned}
\end{equation}

Where $\tau$ refers to Kendall's tau rank correlation metric, $\mathcal{A}_{p}$ and $\mathcal{A}_{r}$ are the vectors of top-1 accuracies of models in the search space under training schemes $p$ and $r$, the proxified and reference training schemes, respectively. $t_p$ is the average training time of the models under evaluation using proxified training scheme $p$, and $t_\text{spec}$ is a desirable upper limit on average training time. The vectors $\mathcal{A}_{p}$ and $\mathcal{A}_{r}$ are of length $n$, which is the number of models under evaluation, and contain the accuracy of corresponding models using the two training schemes. For search, we utilize a uniform grid of $n=20$ models selected based on FLOPs and \# parameters, sampled uniformly from the search space. This grid of models is a representation of the search space owing to an even spread of FLOPs and model size across the search space and encourages the search to find generalizable solutions. A finer grid could be used, but would increase the search cost. We use $t_{spec}=3$ GPU-hours based on available compute.

\begin{figure}[tp!]
\centering
% \vspace{-0.15in}
\includegraphics[width=0.52\linewidth]{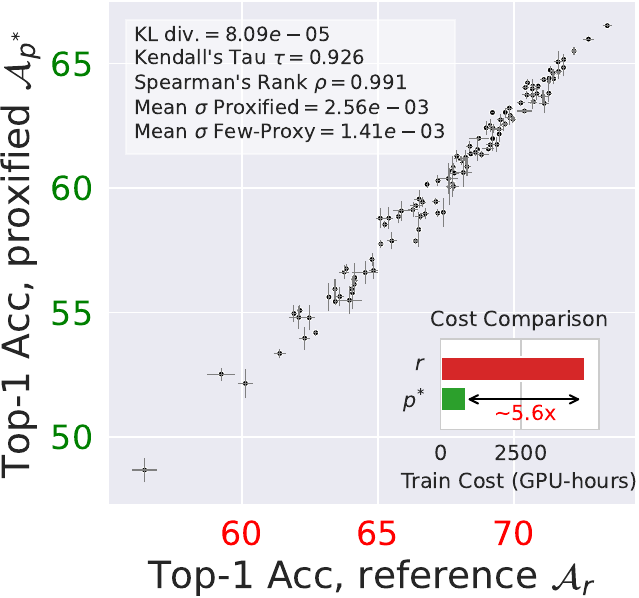}
\vspace{-0.10in}
% \captionsetup{format=plain,justification=justified}
\caption{Validation of $p^{*}$ using 120 random unseen models, trained using both $p^{*}$ and $r$. Architecture rankings are strongly correlated between the two schemes $\tau=0.926$.} \label{fig:abl-underfit}
\vspace{-0.15in}
\end{figure}
The search process is shown in Fig.~\ref{fig:anb-methodology} (top). The goal of the search is to find a proxified training scheme $p=p^{*}$ that satisfies Eq.~\ref{eq:max_tau}. The reference training scheme $r$ is an `ideal' training scheme commonly employed for training models on the dataset being utilized but cannot be used for benchmark construction due to its computational intractability\footnote{We use a scheme from \url{https://github.com/huggingface/pytorch-image-models}}. It is fixed throughout the experiments hence the vector $\mathcal{A}_{r}$ is constant during search. $p$, however, is the proxified training scheme that we optimize for and is parameterized by training hyperparameters $\{b, e_t, e_{s}, e_{f}, res_s, res_f\}$. These hyperparameters pertain to $b$ batch size, $e_t$ total number of training epochs, $e_{s}$ and $e_{f}$ start and finish epoch number for progressive resizing~\cite{karras2017progressive}, and $res_s$ and $res_f$ for start and finish input image resolution for progressive resizing. All six training hyperparameters to optimize are categorical hyperparameters with pre-specified values, and offer a trade-off between convergence speed and accuracy. At every search step, the optimizer selects a $p$ by sampling a value for each of the six hyperparameters. The optimizer then obtains $\mathcal{A}_{p}$ and $t_p$ by training each of the $n$ models using $p$, and computes the rank correlation $\tau$. The resulting $\tau$ and $t_p$ can be used by the optimizer to select better $p$ in future iterations. 

Any viable optimizer can be used to perform this search and is not the focus of this work. We chose to use a fairly trivial grid search owing to the high degree of parallelism that it offers combined with the low dimensionality of the hyperparameter space. Grid search allows the evaluation of multiple proxy configuration $p$ to be performed concurrently, with early stopping when the desired $\tau$ and $t_p$ are achieved. Hence, the search can be massively parallelized over multiple GPU nodes. 

This search results in a training scheme named $p^{*}$ that reduces the training cost by roughly $5.6\times$ (i.e., $\frac{t_{r}}{t_{p^{*}}} \approx 5.6$), while yielding $\tau = 0.94$. We validate the resulting $p^{*}$ by evaluating 120 randomly chosen, previously unseen models from the search space using both $p^{*}$ and $r$ by training and validating each model three times using different seeds. The results in Fig.~\ref{fig:abl-underfit} show the mean proxified accuracies of these 120 models, $\mathcal{A}_{p^{*}}$, on the vertical axis against the mean reference accuracies $\mathcal{A}_{r}$, on the horizontal axis, also plotted are the error bars of the three runs. The validation $\tau=0.926$ shows a strong architecture rank correlation between $p^{*}$ and $r$, showcasing the ability of the resulting training proxy $p^{*}$ to emulate the true architecture ranks while being $5.6\times$ cheaper.

\subsection{Constructing Accel-NASBench}
Having found proxy configuration $p^{*}$ that achieves a strong architecture rank correlation with the reference scheme while being roughly $5.6\times$ cheaper, we collect the datasets for training the surrogate models. Fig.~\ref{fig:anb-methodology} (bottom) shows the dataset collection pipeline.% These datasets consist of architecture-accuracy pairs for accuracy surrogate, and architecture-throughput (latency) pairs for on-device performance surrogates.

\subsubsection{Dataset Collection -- Accuracy.} We collect a dataset of \{architecture, accuracy\} pairs for roughly 5.2k randomly sampled architectures on the ImageNet2012 dataset using the proxy scheme $p^{*}$. We used random sampling as existing works, such as NASBench-301, have shown that unbiased surrogates (i.e., those trained using only randomly sampled data) serve as strong predictors of model performance. We utilized a compute cluster of 6 nodes, each with $4\times$ RTX 3090 GPUs to collect this dataset. This incurs a cost of only 17k GPU-hours. The collected accuracy dataset is named $\texttt{ANB-Acc}$.

\subsubsection{Dataset Collection -- Throughput/Latency.} For training the performance surrogates, we perform on-device measurements of inference throughputs of the 5.2k randomly sampled architectures on 6 accelerator platforms: Google Cloud TPUv2 and TPUv3, NVIDIA RTX 3090 and A100 GPUs, and Xilinx Ultrascale+ ZCU102 and Versal AI Core VCK190 FPGAs. We also performed latency measurements on the FPGA platforms. % These on-device performance datasets would allow the benchmark to be utilized to evaluate search algorithms geared towards bi-objective NAS. 

On \textbf{cloud TPUs}, we measure the inference throughput using Torch/XLA after discarding the TPU warmup phase, which involves XLA graph compilations and caching. We take throughput measurement four times and use the average as the measured value. Similarly, on \textbf{GPU} platforms, we discard warmup throughput measurements and use the mean of two inference runs as the final throughput value of the model. On \textbf{FPGA} platforms, the inference throughput and latency measurements are performed using the Xilinx Vitis AI Deep-Learning Processing Unit (DPU) blocks, which are pre-compiled hardware IPs offered to make the design of FPGA inference accelerators easier. We performed 8-bit post-training quantization of weights for the 5.2k models, followed by cross-compilation for the target DPUs and the platforms. % We measured throughput for the TPU and GPU platforms and both throughput and latency for the FPGAs. 

The collected accelerator performance datasets on the 5.2k architectures are termed \texttt{ANB-\{device\}-\{metric\}}, where throughput (\texttt{Thr}) metric is supported by all devices while latency (\texttt{Lat}) is supported only by FPGAs.

% \begin{table}
%   \begin{minipage}{0.45\linewidth}
%     \centering
%     \caption{Surrogate test performance on \texttt{ANB-Acc}.}
%         \resizebox{\linewidth}{!}{
% 	\begin{tabular}{lccc}
% 		\hline
% 		Model  & $R^2$ & KT $\tau$ & MAE  \\\hline
% 		  XGB & \textbf{0.984} & \textbf{0.922}     & \textbf{3.06e-3} \\
%             LGB & 0.984 & 0.922     & 3.08e-3 \\
% 		  RF        & 0.869 & 0.782 & 8.88e-3 \\
%             $\epsilon$-SVR  & 0.943 & 0.886  & 5.32e-3 \\
%             $\nu$-SVR       & 0.942 & 0.881 & 5.45e-3 \\
%             \hline
% 	\end{tabular}
%          }
% 	\label{tab:fit-anb-acc}
% 	\vspace{-0.1in}
%     % \caption{Table 1}
%   \end{minipage}%
%   \begin{minipage}{0.45\linewidth}
%     \centering
% %%%%%   
%     \caption{XGB test performance on \texttt{ANB-\{device\}-\{metric\}}.}
%     \resizebox{\linewidth}{!}{
% 	\begin{tabular}{lccc}
% 		\hline
% 		Dataset  & $R^2$ & KT $\tau$ & MAE  \\\hline
% 		  \texttt{ANB-ZCU-Thr} & 0.990 & 0.955     & 13.2 \\
%             \texttt{ANB-ZCU-Lat} & 1.000 & 0.987     & 5.2e-2 \\
%             \texttt{ANB-VCK-Thr} & 0.991 & 0.949     & 69.5 \\
%             \texttt{ANB-VCK-Lat} & 0.999 & 0.980     & 4.0e-2 \\
%             \texttt{ANB-TPUv3-Thr} & 0.975 & 0.905     & 29.1 \\
%             \texttt{ANB-TPUv2-Thr} & 0.994 & 0.962     & 14.4 \\
%             \texttt{ANB-A100-Thr} & 0.995 & 0.975     & 159.7 \\
%             \texttt{ANB-RTX-Thr} & 0.996 & 0.968     & 116.1 \\
%             \hline
% 	\end{tabular}
%  }

% %%%%
%     \label{tab:fit-anb-accel}
%   \end{minipage}
% \end{table}
\subsubsection{Surrogate Fitting -- Accel-NASBench}
Having the architecture, accuracy, and on-device performance datasets in hand, we train and compare the predictive performance of a variety of candidate surrogates, including XGBoost, LGBoost, Random Forests, and SVR. %$\epsilon$-SVR, and $\nu$-SVR.

\begin{wraptable}{r}{3.5cm} 
        \vspace{-0.15in}
	\caption{Surrogate test performance on \texttt{ANB-Acc}.}
	\vspace{-0.1in}
	\centering
  \resizebox{3.5cm}{!}{
	\begin{tabular}{lccc}
		\hline
		Model  & $R^2$ & KT $\tau$ & MAE  \\\hline
		  XGB & \textbf{0.984} & \textbf{0.922}     & \textbf{3.06e-3} \\
            LGB & 0.984 & 0.922     & 3.08e-3 \\
		  RF        & 0.869 & 0.782 & 8.88e-3 \\
            $\epsilon$-SVR  & 0.943 & 0.886  & 5.32e-3 \\
            $\nu$-SVR       & 0.942 & 0.881 & 5.45e-3 \\
            \hline
	\end{tabular}
    }
	\label{tab:fit-anb-acc}
	\vspace{-0.1in}
 % \hangindent=0.3em
\end{wraptable}

\begin{figure*}[tp!]
\centering
\includegraphics[width=.88\textwidth]{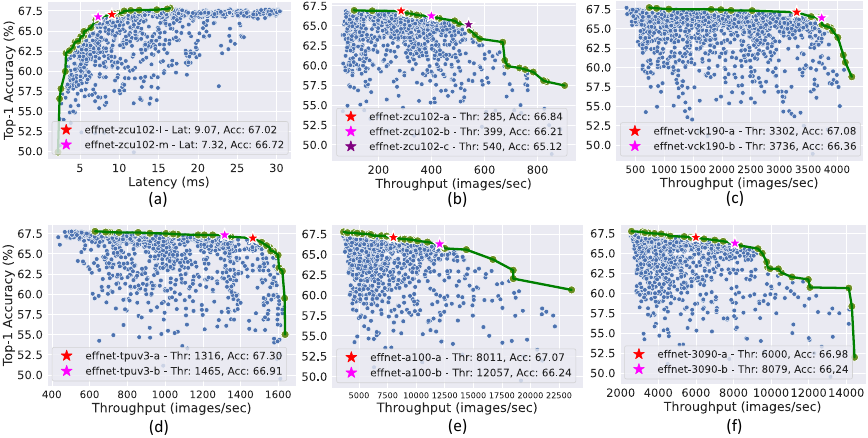}
\vspace{-0.1in}
\caption{Search using RL-based bi-objective optimization. Fig (a) shows the pareto-optimal front using simulated search on accuracy-latency objectives. Fig (b)-(f) show the results of accuracy-throughput search on (b) ZCU102 and (c) VCK190 FPGAs, (d) TPUv3, (e) A100 and (f) RTX 3090 GPUs. Also shown in red, magenta, and purple star markers are pareto-optimal solutions hand-picked for evaluation. Legends show their performances predicted by surrogates.} \label{fig:search-reinforce-rl}
\vspace{-0.1in}
\end{figure*}
We split the collected datasets into train/val/test splits of ratio 0.8/0.1/0.1 and utilize the train/val splits for surrogate hyperparameter tuning. We represent the hyperparameters in ConfigSpace~\citep{lindauer2019boah} and utilize SMAC3~\citep{lindauer2022smac3} for finding the configurations for the surrogates. Finally, we fit the surrogates on the datasets using the train split and evaluate on the test split.

\begin{wraptable}{r}{4cm}
\centering
        \vspace{-0.15in}
        \captionsetup{format=plain,justification=justified}
        \caption{XGB test performance on \texttt{ANB-\{device\}}-\texttt{\{metric\}}.}
	\vspace{-0.1in}
	\centering
 \resizebox{4.0cm}{!}{
	\begin{tabular}{lccc}
		\hline
		Dataset  & $R^2$ & KT $\tau$ & MAE  \\\hline
		  \texttt{ANB-ZCU-Thr} & 0.990 & 0.955     & 13.2 \\
            \texttt{ANB-ZCU-Lat} & 1.000 & 0.987     & 5.2e-2 \\
            \texttt{ANB-VCK-Thr} & 0.991 & 0.949     & 69.5 \\
            \texttt{ANB-VCK-Lat} & 0.999 & 0.980     & 4.0e-2 \\
            \texttt{ANB-TPUv3-Thr} & 0.975 & 0.905     & 29.1 \\
            \texttt{ANB-TPUv2-Thr} & 0.994 & 0.962     & 14.4 \\
            \texttt{ANB-A100-Thr} & 0.995 & 0.975     & 159.7 \\
            \texttt{ANB-RTX-Thr} & 0.996 & 0.968     & 116.1 \\
            \hline
	\end{tabular}
        }
	\label{tab:fit-anb-accel}
	\vspace{-0.1in}
\end{wraptable}
We evaluate the surrogates' fit quality using the coefficient of determination ($R^2$), Kendall's Tau rank correlation $\tau$,  and mean absolute error (MAE). As shown in Table.~\ref{tab:fit-anb-acc}, gradient-boosting techniques outperform all other surrogates in all three evaluation metrics. We also notice a similar trend in the on-device performance surrogates. Table.~\ref{tab:fit-anb-accel} shows the performance of XGBoost on the test datasets of \texttt{ANB-\{device\}-\{metric\}}. %Please note that MAE is measured in images/sec for \texttt{Thr} datasets and milliseconds (ms) for \texttt{Lat} datasets.

\section{Evaluating Accel-NASBench}

\subsection{Evaluating the accuracy surrogate} \label{sec:eval-acc-surr}
We compare the trajectory of search using only the accuracy surrogate against true optimizer runs (using scheme $p^{*}$) on three popular discrete NAS optimizers: Regularized Evolution (RE)~\citep{real2019regularized}, Random Search (RS)~\citep{li2020random}, and REINFORCE~\citep{zoph2016neural}. The trajectories using simulated runs (i.e., surrogate-based) are averaged over five runs with different random seeds; however, the true runs are only performed once owing to the high evaluation cost of each run (similar to~\cite{siems2020bench}). %(as also performed by~\citet{siems2020bench}). 

\begin{figure}[t]
\centering
\includegraphics[width=.98\linewidth]{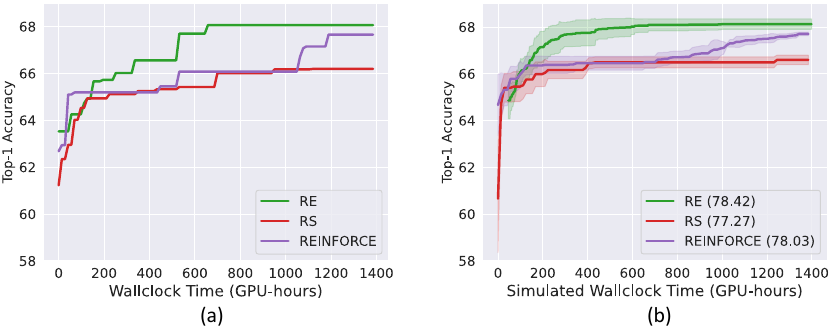}
\vspace{-0.15in}
\caption{Comparison of trajectory of uni-objective search between (a) true and (b) simulated runs using Accel-NASBench.}
%Also shown in the legend of (b) are the true top-1 ImageNet accuracy of best models found from each of the three search methods.
\vspace{-0.20in}
\label{fig:eval-trajectory}
\end{figure}
\subsubsection{Results.} Fig.~\ref{fig:eval-trajectory} shows the trajectory of (a) true and (b) simulated search. The accuracy surrogate is able to mimic the behaviour of true search as shown by the similarity in trajectories, with RS predictably underperforming compared to REINFORCE and RE owing to the high variability of model performance in the search space. This is in contrast to the dense and highly predictable DARTS search space in which RS is able to achieve near state-of-the-art performance~\citep{li2020random}. On MnasNet space, RS stagnates fairly early while RE and REINFORCE achieve significantly better results. This is in line with the findings of existing works (such as~\citet{siems2020bench}).

\begin{figure*}[tp!]
\centering
\includegraphics[width=0.8\textwidth]{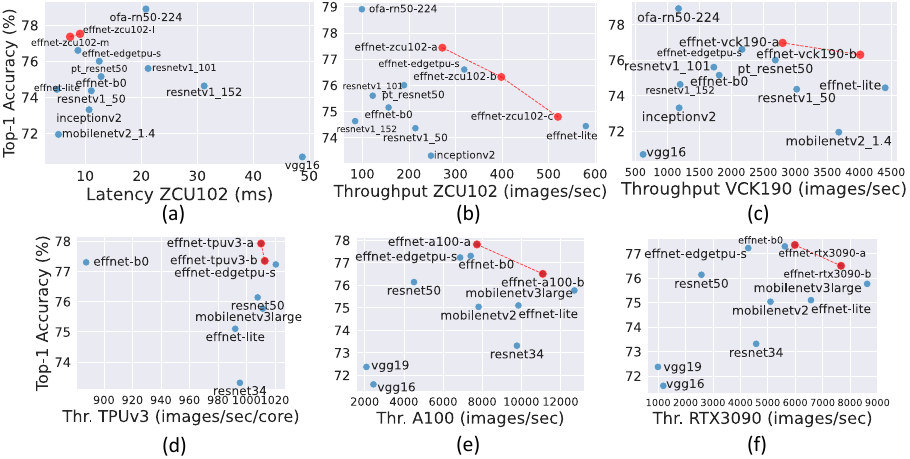}
\vspace{-0.12in}
\caption{Evaluation true results comparison against high-quality models on (a) accuracy-latency tradeoff on ZCU102, and accuracy-throughput tradeoff on (b) ZCU102, (c) VCK190 FPGAs, (d) TPUv3, (e) A100 and (f) RTX-3090 GPUs. In red markers are the evaluation results of our search using accuracy-throughput/latency bi-objective search using REINFORCE. The zero-cost search using Accel-NASBench yields models that compare to known high-quality solutions in the MnasNet search space (e.g., EfficientNet-b0 and EfficientNet-EdgeTPU-S). Existing FPGA results obtained from Vitis-AI model zoo.} \label{fig:eval-results}
\vspace{-0.15in}
\end{figure*}

% \vspace{-0.9\baselineskip}
\subsection{Accel-NASBench for bi-objective search} \label{sec:eval-biobj}
We evaluate the performance of Accel-NASBench for bi-objective accuracy-throughput (latency) RL-based search using REINFORCE \cite{zoph2016neural}. Towards this end, instead of comparing the search trajectory between surrogates-based search and true search, we show that the search results using surrogates yield models that are comparable in accuracy and performance to those found by existing works that perform true search~\citep{tan2019efficientnet, gupta2020accelerator}. We do this to avoid the exorbitant compute cost and complex instrumentation of true bi-objective discrete RL-based search, which would require both accuracy and on-device throughput measurements in a single, continuous pipeline. Having not utilized any \textit{model} or \textit{dataset} proxies in the construction of Accel-NASBench, the search results can be compared directly against known high-quality models (such as~\citet{tan2019efficientnet, gupta2020accelerator}). We perform searches using the throughput surrogates of five devices and the latency surrogate of ZCU102 FPGA. Fig.~\ref{fig:search-reinforce-rl} (a) shows the pareto-optimal solutions resulting from accuracy-latency bi-objective search using the ZCU102 FPGA latency surrogate. Fig.~\ref{fig:search-reinforce-rl} (b)-(f) shows the accuracy-throughput search results targeting five accelerators. 

\textbf{Evaluation.} Since the accuracy surrogate predicts the accuracy under the proxified training scheme ($p^{*}$) rather than the true accuracy, we evaluate a few hand-picked pareto-optimal solutions from the searches by training using the reference training scheme $r$, and by performing on-device throughput/latency measurements. The evaluation true accuracy and throughput/latency results are plotted in Fig.~\ref{fig:eval-results} for the 5 hardware platforms. We compare the obtained accuracy-throughput against existing state-of-the-art searched and handcrafted models. Results show performance improvements compared to accuracy-FLOPs optimization works such as EfficientNet-B0~\citep{tan2019efficientnet} and MobileNetsV3~\citep{howard2019searching}. For example, our \texttt{effnet-vck190-a} achieves $1.8\%$ higher accuracy and $55.0\%$ better throughput than \texttt{effnet-b0}~\citep{tan2019efficientnet}, and $0.37\%$ higher accuracy and $29.4\%$ better throughput than \texttt{effnet-edgetpu-s}~\citep{gupta2020accelerator}, on the VCK190 FPGA platform. %Please note that on the FPGA platforms, Figure.~\ref{fig:eval-results} (a)-(b) compares the 8-bit quantized accuracies of the models since the standard FPGA deployment flow involves quantizing weights and inputs to 8-bits.

Given the fact that surrogate-based zero-cost search is able to find models that offer comparable accuracy-throughput trade-offs against existing high-quality true search results, we conjecture that the surrogates' predictive distributions mimic the true accuracy/throughput distributions fairly accurately since the NAS optimizer is able to effectively explore the well-performing regions of the search space. 
\vspace{-0.10in}
\section{Conclusion}
We presented an approach that allows for computationally efficient construction of NAS benchmarks for large-scale datasets. The presented approach searches for training proxies that maintain architecture rankings relative to their true ranks while reducing model training costs. Using the proposed approach, we built Accel-NASBench, the first NAS benchmark for the ImageNet dataset, which includes hardware accelerator performance benchmarks for GPUs, TPUs, and FPGAs. We validated Accel-NASBench using a suite of experiments with various NAS optimizers, hardware platforms, and optimization objectives. We hope that our approach will guide the development of more high-quality benchmarks for large-scale datasets, and Accel-NASBench will serve as a useful zero-cost tool for NAS researchers in evaluating their techniques.
\vspace{-0.10in}

% \begin{acks}

% \end{acks}

\bibliographystyle{ACM-Reference-Format}
\bibliography{bibliography}


\end{document}